\documentclass[journal]{IEEEtran}

\usepackage{graphics,subfigure} 
\usepackage{epsfig} 
\usepackage{mathptmx} 
\usepackage{times} 
\usepackage{amsmath} 
\usepackage{amssymb}  
\usepackage{captcont}
\usepackage{graphicx}
\usepackage{multirow}
\usepackage{calc}
\usepackage{tikz}
\usepackage{algorithm}
\usepackage{algorithmic}
\usepackage{booktabs}
\usepackage{flushend}
\usepackage{slashbox}
\usepackage{cases}
\usepackage{array}
\usepackage{makecell}
\usepackage{bbm}
\usepackage{bm}
\usepackage{cite}
\ifCLASSINFOpdf
\else
\fi
\begin{document}
\title{\huge Robust Multi-class Feature Selection via $l_{2,0}$-Norm Regularization Minimization}

\author{Zhenzhen~Sun,
        and Yuanlong~Yu,
\thanks{Z. Sun and Y. Yu are with the College of Mathematics and Computer Science Fuzhou University, Fuzhou, Fujian, 350116, China.}
\thanks{Corresponding author: Y. Yu (email: yu.yuanlong@fzu.edu.cn.)}.
\thanks{This work is supported by National Natural Science Foundation of China (NSFC) under grant $\#61873067$.}}
\markboth{}%
{Zhen \MakeLowercase{\textit{et al.}}: Bare Demo of IEEEtran.cls for IEEE Journals}
\maketitle
\begin{abstract}
Feature selection is an important data preprocessing in data mining and machine learning, which can reduce feature size without deteriorating model's performance. Recently, sparse regression based feature selection methods have received considerable attention due to their good performance. However, because the $l_{2,0}$-norm regularization term is non-convex, this problem is very hard to solve. In this paper, unlike most of the other methods which only solve the approximate problem, a novel method based on homotopy iterative hard threshold (HIHT) is proposed to solve the $l_{2,0}$-norm regularization least square problem directly for multi-class feature selection, which can produce exact row-sparsity solution for the weights matrix. What'more, in order to reduce the computational time of HIHT, an acceleration version of HIHT (AHIHT) is derived. Extensive experiments on eight biological datasets show that the proposed method can achieve higher classification accuracy (ACC) with fewer number of selected features (No.fea) comparing with the approximate convex counterparts and other state-of-the-art feature selection methods. The robustness of classification accuracy to the regularization factor and the number of selected feature are also exhibited.
\end{abstract}


\section{Introduction}\label{Intro}
Feature selection, the process of selecting a subset of features which are the most relevant and informative, has been widely researched for many years \cite{Li:14,Zhao:16,Suchetha:19,Zabihimayvan:19,Qasim:20}. Feature selection has become an essential component in data mining and machine learning because it can reduce the feature size, enhance data understanding, alleviate the effect of the curse of dimensionality, speed up the learning process and improve model's performance. Therefore, it has been widely used in many real-world applications, e.g., text mining \cite{Tang:16,Wan:19}, pattern recognition \cite{Suchetha:19}, and bioinformatics \cite{Saeys:07,Lei:20}.

In general, feature selection methods can be divided into three categories depending on how they combine the feature selection search with model learning algorithms: filter methods, wrapper methods, and embedded methods. In the filter methods, features are selected according to the intrinsic properties of the data before running learning algorithm. Therefore, filter methods are independent of the learning algorithms and can be characterized by utilizing the statistical information. Typical filter methods include Relief \cite{Kira:92}, Chi-square and information gain \cite{Raileanu:04}, mRMR \cite{Peng:05}, \emph{etc}. The wrapper methods use learning algorithm as a black box to score subsets of features, such as correlation-based feature selection (CFS) \cite{Hall:99} and support vector machine recursive feature elimination (SVM-RFE) \cite{Guyon:02}. Embedded methods incorporate the feature selection and model learning into a single optimization problem, such that higher computational efficiency and classification performance can be gained than filter methods and wrapped methods. Thus, the embedded methods have attracted large attention these years.

Recently, with the development of sparsity researches, sparsity regularization has been widely applied in embedded feature selection methods. The concern behind this is that selecting a minority of features is naturally a problem with sparsity. For binary classification task, the feature selection task can be tackled by $l_0$-norm minimization \cite{Weston:03} directly in which features corresponding to non-zero weights are selected. However, the non-convexity and non-smoothness of $l_0$-norm make it very hard to solve. Most methods relax the $l_0$-norm by $l_1$-norm to make the minimization problem be convex and easy to solve, which is called LASSO \cite{Tibshirani:96}. Although some strategies such as one-versus-one or one-versus-all can be used to expand LASSO for multi-class feature selection problem, structural sparsity models are more desirable so that we can obtain the shared pattern of sparsity. Inspired by that, lots of methods have been proposed based on structural sparsity for multi-class feature selection \cite{Gui:17}. In \cite{Nie:10}, Nie \emph{et al.} proposed a robust feature selection (RFS) method with emphasizing joint $l_{2,1}$-norm minimization on both loss function and regularization. After that, $l_{2,1}$-norm has been widely used in multi-class feature selection methods, e.g., UDFS \cite{Yang:11}, L-FS \cite{Zhang:15}, RLSR \cite{Chen:17}, URAFS \cite{Li:19}, \emph{etc}.

Though satisfactory results can been achieved by using $l_{2,1}$-norm regularization for multi-class feature selection, there are some limitations. First, $l_{2,1}$-norm is just an approximation of $l_{2,0}$-norm, thus the solutions are essentially different from the original optimum value. Second, Qian \emph{et al.} \cite{Qian:15} proved that $l_{2,1}$-norm over-penalizes features with large weights, which will lead to an unfair competition between different features and hurt data approximation performance. Last, it is hard to tune the regularization parameter of $l_{2,1}$-norm to get exact row-sparsity solution, even a large regularization factor (e.g., $10^5$) cannot produce strong row-sparsity. Thus it is not clear how many features will be selected if we tune the regularization parameter. Consequently, it is significant to find a method to solve the original $l_{2,0}$-norm regularization problem.

In \cite{Cai:13}, Cai \emph{et al.} proposed a robust and pragmatic multi-class feature selection (RPMFS) method based on $l_{2,0}$-norm equality constrained optimization problem. RPMFS sets the objective function as a $l_{2,1}$-norm loss term with a $l_{2,0}$-norm equality constraint, and uses the augmented Lagrangian method to solve this equality constrained problem. In \cite{Pang:19}, Pang \emph{et al.} also propose an efficient sparse feature selection method (ESFS) based on $l_{2,0}$-norm equality constrained problem, then they transform the model into the same structure as LDA to calculate the ratio of inter-class scatter to intra-class scatter of features. However, the $l_{2,0}$-norm equality constrained based methods exist two defects. First, the number of selected features (No.fea) need to be pre-defined to construct the equality constraint, and once the No.fea is re-tuned, the program must be re-run, which will cost a large amount of time and effort and is unpractical. Second, the overall performances of these methods are very sensitive to No.fea, in order to obtain satisfactory classification results, the number of selected features need to be tuned carefully. From literatures of sparsity research, it has been proven that the regularized problem is more effective than the equality constraint problem to find a sparse solution.

Therefore, this paper  proposes a novel and simple framework for multi-class feature selection which solve the original $l_{2,0}$-norm regularization least square problem (denoted as $l_{2,0}$-FS) directly. In order to effectively solve the proposed objective function, the homotopy iterative hard threshold method (HIHT) is introduced to perform the optimization. What'more, an acceleration version of HIHT (AHIHT) is derived to reduce the computational time. After learning, an exact row-sparsity solution is produced and the features can be selected in group. The contribution of this proposed $l_{2,0}$-FS is that an effective and efficient optimization algorithm is designed to solve the $l_{2,0}$-norm regularization least square problem and produce row-sparsity solution for feature selection. After learning, the number of selected features can be tuned quickly without rerunning the optimization program. To evaluate the performance of the proposed method, we use the selected features for classification, and compare the results with Baseline (without feature selection) and six supervised feature selection methods in terms of classification accuracy (ACC) and the number of selected features (No.fea) over eight biological data sets. The results show that the features selected by $l_{2,0}$-FS are superior to those selected by the comparison methods.

The rest of this paper is organized as follows. Section \ref{RelW} presents the notations and definitions used in this paper, and related works on $l_{2,1}$-norm regularized problem and $l_{2,0}$-norm equality constrained problem based multi-class feature selection methods are introduced. In Section \ref{Proposed}, the proposed method $l_{2,0}$-FS is presented and the optimization algorithms are introduced. The experimental results are presented in Section \ref{Exp}. Conclusions and future work are given in Section \ref{Con}.
\section{Related Work}\label{RelW}
\subsection{Notations and Definition}
The notations and definitions used in this paper are shown in this subsection. Vectors are written as boldface lowercase letters and matrices are written as boldface uppercase letters. For a vector $ \textbf{x} \in R^n $, $x_i$ denotes the $i$-th element of $\textbf{x}$. For a matrix $ \textbf{X}=\{x_{ij}\}  \in R^{n\times m} $, $ \textbf{x}^{i} $ and $ \textbf{x}_{j}$ denote its $ i$-th row and $j$-th column, respectively.

For $p \ne 0$, the $p$-norm of the vector $ \textbf{x}$ is defined as:
\begin{equation}\notag
||\textbf{x}||_p=(\sum_{i=1}^n{|x_i|^p}) ^{\text{1/}p}.
\end{equation}

The $l_0$-norm of the vector $ \textbf{x}$ is defined as:
\begin{equation}\notag
||\textbf{x}||_0 = \ \sum_{i=1}^n{|x_i|^0},
\end{equation}
which count the number of non-zero elements in $\textbf{x}$.

The Frobenius norm of $ \textbf{X} $ is defined as:
\begin{equation}\notag
||\textbf{X}||_F = \ \sqrt{\sum_{i = 1}^n{\sum_{j = 1}^m{x_{ij}^2}}} = \ \sqrt{tr(\textbf{X}^T \textbf{X})}. \\
\end{equation}

The $l_{2,1}$-norm of $ \textbf{X} $ is defined as:
\begin{equation}\notag
||\textbf{X}||_{2,1} =  \ \sum_{i =1}^n{||\textbf{x}^i||_2} = \ \sum_{i =1}^n{\sqrt{\sum_{j = 1}^m{x_{ij}^2}}}. \\
\end{equation}

The $l_{2,0}$-norm of matrix $ \textbf{X} $ is defined as:
\begin{equation}\notag
||\textbf{X}||_{2,0} =  \ \sum_{i =1}^n{ \mathbbm{1}_{||\textbf{x}^{i}||_2 \ne 0}}, \\
\end{equation}
where $ \mathbbm{1}_{A} $ stands for the indicator function. For a scalar $x$, if $x \ne 0$, $\mathbbm{1}_{x} = 1$, otherwise $\mathbbm{1}_{x} = 0$. Thus the $l_{2,0}$-norm of matrix $ \textbf{X} $ is defined as the number of non-zero rows in $ \textbf{X} $. If a matrix has a large number of zero rows, we say it has the property of row-sparsity.
\subsection{Multi-class Feature Selection based on $l_{2,1}$-Norm}\label{l21-norm methods}
In general, most multi-class feature selection algorithms based on $l_{2,1}$-norm regularization can be formulated as follows:
\begin{equation}\label{l-21}
  \underset{\textbf{W,b}}{\min} \ ||\textbf{W}^T\textbf{X}+\textbf{b}\textbf{1}^T-\textbf{Y}||_{F}^{2}+\lambda ||\textbf{W}||_{2,1},
\end{equation}
where $\textbf{X}=\{\textbf{x}_1, \textbf{x}_2,...,\textbf{x}_N\} \in R^{d \times N}$ is the training data. $\textbf{Y}=\{\textbf{y}_1, \textbf{y}_2,...,\textbf{y}_N\} \in B^{C \times N}$ is the binary label matrix with $y_{ij} = 1$ if $\textbf{x}_i$ has label $y_i=j$;
otherwise $y_{ij} = 0$. $\textbf{W} \in R^{d \times C}$ denotes the model weights and  $\textbf{b} \in R^{C \times 1}$ denotes the learned biased vector. $\textbf{1} \in R^{N \times 1}$ is a column vector with all its entries being $1$. $N$ is the sample number, $d$ is the feature dimension, $C$ denotes the class number, and $\lambda$ is the regularization factor. After optimizing, the features of $\textbf{X}$ are selected according to the magnitude of $||\textbf{w}^i||_2$.

This problem has been widely studied and a lot of variants have been proposed. Nie \emph{et al.} \cite{Nie:10} first combined the $l_{2,1}$-norm regularization with a $l_{2,1}$-norm loss term instead of the Frobenius norm loss term and demonstrated the proposed model is more robust for outliers than the original model. Yang \emph{et al.} \cite{Yang:11} incorporated discriminative analysis into the $l_{2,1}$-norm minimization. By doing that, an unsupervised feature selection joint model was yielded. In \cite{Zhang:15}, the authors combine the $l_{2,1}$-norm regularization with the fisher criterion to select more discriminative features. recently, Yan \emph{et al.} \cite{Yan:16} imposed both nonnegative and $l_{2,1}$-norm constraints on the feature weights matrix. The nonnegative property ensures the row-sparsity of learned feature weights combining with the $l_{2,1}$-norm minimization, which makes it clearer for which feature should be selected.

Although the $l_{2,1}$-norm based model can achieve satisfactory results, one of the biggest problem is that we don't know how many features need to be selected after the regularization factor tuning. From the sparsity perspective, $l_{2,0}$-norm is more desirable.
\subsection{Multi-Class Feature Selection based on $l_{2,0}$-Norm Constrained Problem}\label{l20-norm methods}
The multi-class feature selection based on $l_{2,0}$-norm always construct an equality constraint to determine the number of non-zero rows of weights matrix. In \cite{Cai:13}, Cai \emph{et al.} construct the objective function as a $l_{2,1}$-norm loss term with a $l_{2,0}$-norm equality constraint, which can be written as follows:
\begin{align}\label{RPMFS}
\underset{\textbf{W,b}}{\min}\,\,||\textbf{W}^T\textbf{X}+\textbf{b}\textbf{1}^T-\textbf{Y}||_{\text{2,}1}
\notag \\
s.t. ||\textbf{W}||_{\text{2,}0}=k, \ \ \ \ \ \ \ \ \ \ \ \
\end{align}
then they used the Augmented Lagrangian Multiplier (ALM) method to solve this problem.

In \cite{Pang:19}, the authors form a similar model which is written as follows:
\begin{align}\label{ESFS}
\underset{\textbf{W,b}}{\min}\,\,||\textbf{W}^T\textbf{X}+\textbf{b}\textbf{1}^T-\textbf{YQ}||_{\text{2,}1}
\notag \\
s.t. ||\textbf{W}||_{\text{2,}0}=k, \ \ \ \ \ \ \ \ \ \ \ \
\end{align}
where $ \textbf{Q} \in R^{C \times C}$ can be any reversible matrix which is used to code labels. By using
this label coding method, they transform model \eqref{ESFS} into the same structure as LDA which can calculate the ratio of inter-class scatter to intra-class scatter of features.

Though the above mentioned feature selection methods are based on $l_{2,0}$-norm, they also need to predefined the number of selected features to construct the equality constraint. How many features need be selected is unknown, so it will cost a large amount of time and effort to tune the number of selected features to get a satisfactory result. From literatures of sparsity research, it has been proven that the regularized problem is more effective than the equality constraint problem to find a sparse solution.
\section{The Proposed method}\label{Proposed}
\subsection{Problem Formulation}
In this work, we propose a novel multi-class feature selection method by using the least square regression combined with a $l_{2,0}$-norm regularization, the optimization function is formulated as follows:
\begin{equation}\label{l-20}
  \varphi_{\lambda} (\textbf{W}) = \underset{\textbf{W}}{\min} \ \frac{1}{2}||\textbf{W}^T\textbf{X}+\textbf{b}\textbf{1}^T-\textbf{Y}||_{F}^{2}+\lambda ||\textbf{W}||_{2,0},
\end{equation}
where the notations are defined as in \ref{l21-norm methods}.

According to Karush-Kuhn-Tucker (KKT) theorem \cite{Karush:39}, set the derivative of \eqref{l-20} with respect to $\textbf{b}$ to $0$, then we can obtain the optimal solution of $\textbf{b}$ as follows:
\begin{equation}\label{solutionofb}
  \mathbf{b}=\frac{1}{N}\left( \mathbf{Y}-\mathbf{W}^T\mathbf{X} \right) \mathbf{1}.
\end{equation}
Replacing $\textbf{b}$ with equation.~\eqref{solutionofb}, we can rewrite problem \eqref{l-20} as:
\begin{equation}\label{l-201}
 \varphi_{\lambda} (\textbf{W})=\underset{\mathbf{W}}{\min}\,\,\frac{1}{2}||(\mathbf{W}^T\mathbf{X}-\mathbf{Y})\mathbf{H}||_{F}^{2}+\lambda ||\mathbf{W||}_{\text{2,}0},
\end{equation}
where $\textbf{H}=\textbf{I}-\left( \text{1/}N \right) \mathbf{11}^T$ is referred as the centering matrix, which has following properties:
\begin{equation}\label{H}
  \mathbf{H}=\mathbf{H}^T=\mathbf{HH}^T=\mathbf{H}^T\mathbf{H}.
\end{equation}

Now, the objective function of $l_{2,0}$-FS is presented totally, next we will show how to optimize it. Iterative hard threshold (IHT) algorithm \cite{Blumensath:08,Blumensath:09,Lu:14,Jiang:18} is usually used to solve $l_0$-norm regularization minimization for vector sparsity problem, inspired by it, we extended it to solve the matrix sparsity problem \eqref{l-201}.
\subsection{Optimization Algorithm}
Denotes $\mathbf{\tilde{X}}=\mathbf{XH}$ and $\mathbf{\tilde{Y}}=\textbf{YH}$, \eqref{l-201} can be rewritten as:
\begin{equation}\label{l-202}
  \varphi _{\lambda}(\mathbf{W}) =\underset{\mathbf{W}}{\min}\,\,\frac{1}{2}||\mathbf{W}^T\mathbf{\tilde{X}}-\mathbf{\tilde{Y}||}_{F}^{2}+\lambda ||\mathbf{W||}_{2,0}.
\end{equation}

The core idea of the IHT algorithm is using the proximal point technique to iteratively update the solution. Firstly we define $f(\textbf{W})$ as
\begin{equation}\label{LSR}
f(\textbf{W}) = \frac{1}{2}||\textbf{W}^T\mathbf{\tilde{X}} - \mathbf{\tilde{Y}}||_F^2.
\end{equation}

Since $f(\textbf{\textbf{W}}) $ is a differentiable convex function whose gradient is Lipschitz continuous (denote its Lipschitz constant as $L_f$), it can be approximatively iteratively updated by the projected gradient method:
\begin{align}\label{PG}
\textbf{W}^{t+1}  &=  \arg \mathop {\min }\limits_\textbf{W} f(\textbf{W}^{t} ) + tr(\nabla f(\textbf{W}^{t} )^T \left(\textbf{W}  - \textbf{W}^{t} \right)) + \frac{{L}}{2}||\textbf{W}  - \textbf{W}^{t} ||_F^2,
\end{align}
where $ L \ge 0 $ is a constant, which should essentially be an upper bound on the Lipschitz constant of $\nabla f(\textbf{W})$, i.e., $L \geqslant L_f$.

By adding $\lambda ||\textbf{W}||_{2,0}$ into both sides of \eqref{PG}, the solution of \eqref{l-202} can be obtained by iteratively solving the subproblem:
\begin{align}\label{IHT}
\textbf{W}^{t+1}  &=  \arg \mathop {\min }\limits_\textbf{W} f(\textbf{W}^{t} ) + tr(\nabla f(\textbf{W}^{t} )^T \left(\textbf{W}  - \textbf{W}^{t} \right)) \notag \\
& \ \ \ \ + \frac{{L}}{2}||\textbf{W}  - \textbf{W}^{t} ||_F^2 + \lambda ||\textbf{W}||_{2,0}.
\end{align}

By removing the item $  f(\textbf{W}^t) $ and adding an item $ \frac{1}{{2L }}|| \nabla f(\textbf{W}^t )||_F^2 $, both of which are independent on $ \textbf{W} $ and can be considered as constant items, the right hand of \eqref{IHT} can be rewritten as:
\begin{align}\label{IHT1}
\textbf{W}^{t + 1}= \arg \mathop {\min }\limits_\textbf{W}  \frac{L}{2}\{ ||\textbf{W} - (\textbf{W}^t  - \frac{1}{{L}}\nabla f(\textbf{W}^t )||_F^2 + \frac{{2\lambda }}{{L}}||\textbf{W}||_{2,0} \},
\end{align}

Since the Frobenius norm and $l_{2,0}$-norm are all separable function, each row of $\textbf{W}$ can be updated individually:
\begin{align}\label{IHT2}
(\textbf{w}^i) ^{t+1} &= \arg \mathop {\min }\limits_{\textbf{w}^i} \frac{L}{2}\{||\textbf{w}^i-(( \textbf{w}^i)^t-\frac{1}{L}\nabla f((\textbf{w}^i)^t)) ||_{2}^{2}
\notag  \\
 & \ \ \ \ \ +\frac{2\lambda}{L}\mathbbm{1}_{||\textbf{w}^i||_2\ne 0} \},
\end{align}
this subproblem has a closed-form solution as:
\begin{align}\label{eq:hiht_solution}
(\textbf{w}^i)^{t+1}  = \left\{ {\begin{array}{*{20}c}
	{(\textbf{w}^i)^{t} - \frac{1}{{L}}\nabla f((\textbf{w}^i)^{t}),if\;||(\textbf{w}^i)^{t}  - \frac{\nabla f((\textbf{w}^i)^{t})}{{L}}||_2^2  > \frac{{2\lambda }}{{L}}}  \\
	{\textbf{0},\;\;otherwise\;\;\;\;\;\;\;\;\;\;\;\;\;\;\;\;\;\;\;\;\;\;\;\;\;\;\;\;}  \\
	\end{array}} \right.
\end{align}

In \eqref{eq:hiht_solution}, it can be seen that there are a parameter need to be tuned: $L$. The upper bound on $L_f$ is unknown or may not be easily calculated, thus we use line search method to search $L$ as suggested in \cite{Lu:14} until the objective value descent.

\emph{\textbf{Homotopy Strategy}}: many works \cite{Figueiredo:07,Dong:18,Ge:19} have verified that the sparse coding approaches benefit from a good starting point. Thus we can use the solution of \eqref{l-202}, for a given value of $\lambda$, to initialize IHT in a nearby value of $\lambda$. Usually, the next loop with warm starting will require fewer iterations than current loop. Using this warm-starting technique, we can efficiently solve for a sequence of values of $\lambda$, which is called homotopy strategy. An outline of the proposed HIHT algorithm for solving \eqref{l-202} is described as Alg.~\ref{Alg:HIHT}.
\begin{algorithm}[t]
\caption{Homotopy iterative hard threshold method to solve problem \eqref{l-20} }
\label{Alg:HIHT}
\begin{algorithmic}
  \STATE \text{\textbf{(Input:)} Training data $\textbf{X} \in R^{d \times N}$, training labels $\textbf{Y} \in R^{C \times N} $,}
  \STATE \text{centering matrix $\textbf{H} \in R^{N \times N}$, parameters $L_0,{\lambda}_0;$}
  \STATE \text{\textbf{(Output:)} \ $\textbf{W}^*$;}
  \STATE \text{1: initialize $k \gets 0, \rho \in (0,1),\gamma >1, \eta >0, \epsilon >0, \textbf{W}^0=\textbf{0}$;} \\
          \text{2: \textbf{repeat}}\\
          \text{3: \ \ $i \gets 0$;}\\
          \text{4: \ \ $\textbf{W}^{k,0}=\textbf{W}^k$;}\\
          \text{5: \ \ $L_{k,0} \gets L_k$;}\\
          \text{6: \ \ \textbf{repeat}}\\
          \text{ \ \ \ \ \emph{An L-tuning iteration indexed by $i$}}\\
          \text{7: \ \ \ \ update $\textbf{W}^{k,i+1}$ by Eq.~\eqref{eq:hiht_solution};}\\
          \text{8: \ \ \ \ \textbf{while} \ $ {\varphi}_{\lambda _k}(\textbf{W}^{k,i})- {\varphi}_{\lambda _k}(\textbf{W}^{k,i+1})< \frac{\eta}{2} ||\textbf{W}^{k,i}-\textbf{W}^{k,i+1}||_F^2$ \ \textbf{do}}\\
          \text{9: \ \ \ \ \ \ $L_{k,i} \gets \gamma L_{k,i}$;}\\
          \text{10: \ \ \ \ \ update $\textbf{W}^{k,i+1}$ by Eq.~\eqref{eq:hiht_solution};}\\
          \text{11: \ \ \ \ \textbf{end while}}\\
          \text{12: \ \ \ \ $L_{k,i+1} \gets L_{k,i}$;}\\
          \text{13: \ \ \ \ $i \gets i+1$;}\\
          \text{14: \ \ \textbf{until}$||\textbf{W}^{k,i}-\textbf{W}^{k,i+1}||^2_F \le \epsilon$}\\
          \text{15: \ \ $\textbf{W}^{k+1} \gets \textbf{W}^{k,i}$;}\\
          \text{16: \ \ $L_{k+1} \gets L_{k,i}$.}\\
          \text{17: \ \ ${\lambda}_{k+1} \gets \rho {\lambda}_k$;}\\
          \text{18: \ \ $k \gets k+1$;}\\
          \text{19: \textbf{until} ${\lambda}_{k+1}$ is small enough}\\
          \text{20: $\textbf{W}^{\star} \gets \textbf{W}^k$.}
\end{algorithmic}
\end{algorithm}
\subsection{Convergence Analysis}
For a fixed $\lambda_t$, since $\bigtriangledown f$ is Lipschitz continuous with constant $L_f$, we have:
\begin{equation}\label{EQ2}
  \begin{split}
    f(\textbf{W}^{k+1}) \le f(\textbf{W}^k) + tr(\bigtriangledown f(\textbf{W}^k)^T (\textbf{W}^{k+1}-\textbf{W}^k))\\
    + \frac{L_f}{2} ||\textbf{W}^{k+1}-\textbf{W}^k||^2. \ \ \ \ \ \ \ \ \ \ \ \ \ \ \ \ \
\end{split}
\end{equation}

Using this inequality, the fact that $L>L_f$, and \eqref{IHT}, we obtain that
\begin{equation}\label{EQ3}
   \begin{split}
      {\varphi}_{\lambda_t}(\textbf{W}^{k+1})=f(\textbf{W}^{k+1})+ \lambda_t ||\textbf{W}^{k+1}||_0\ \ \ \ \ \ \ \ \ \ \ \ \ \ \ \ \ \ \ \ \ \ \ \ \ \ \ \ \ \ \ \ \ \ \ \ \ \\
        \le f(\textbf{W}^k) + tr(\bigtriangledown f(\textbf{W}^k)^T (\textbf{W}^{k+1}-\textbf{W}^k))+ \frac{L_f}{2} ||\textbf{W}^{k+1}-\textbf{W}^k||^2 \ \ \ \ \ \ \ \ \\
         + \lambda_t ||\textbf{W}^{k+1}||_0 \ \ \ \ \ \ \ \ \ \ \ \ \ \ \ \ \ \ \ \ \ \ \ \ \ \ \ \ \ \ \ \ \ \ \ \ \ \ \ \ \ \ \ \ \ \ \ \ \ \ \ \ \ \ \ \ \ \\
        \le f(\textbf{W}^k) + tr(\bigtriangledown f(\textbf{W}^k)^T (\textbf{W}^{k+1}-\textbf{W}^k))+ \frac{L}{2} ||\textbf{W}^{k+1}-\textbf{W}^k||^2 \ \ \ \ \ \ \ \ \ \\
         + \lambda_t ||\textbf{W}^{k+1}||_0  \ \ \ \ \ \ \ \ \ \ \ \ \ \ \ \ \ \ \ \ \ \ \ \ \ \ \ \ \ \ \ \ \ \ \ \ \ \ \ \ \ \ \ \ \ \ \  \ \ \ \ \ \ \ \ \ \  \\
        \le f(\textbf{W}^k)+ \lambda_t ||\textbf{W}^k||_0 = {\varphi}_{\lambda_t}(\textbf{W}^k),\ \ \ \ \ \ \ \ \ \ \ \ \ \ \ \ \ \ \ \ \ \ \ \ \ \ \ \ \ \ \ \ \ \ \ \ \ \ \ \
   \end{split}
\end{equation}
where the last inequality follows from \eqref{IHT}. The above inequality implies that for a  fixed $\lambda_t$, ${\varphi}_{\lambda_t}\{\textbf{W}^k\}$ is non-increasing and moreover,
\begin{equation}\label{EQ4}
{\varphi}_{\lambda_t}(\textbf{W}^k)-{\varphi}_{\lambda_t}(\textbf{W}^{k+1}) \ge \frac{L - L_f}{2} ||\textbf{W}^{k+1} - \textbf{W}^k||^2.
\end{equation}

Since $f(\textbf{W})$ is bounded below, it then follows that ${\varphi}_{\lambda_t}\{\textbf{W}^k\}$ is bounded below. Hence, ${\varphi}_{\lambda_t}\{\textbf{W}^k\}$ converges to a finite value as $k\rightarrow \infty $ and a local optimal solution $\textbf{W}_{\lambda_t}^*$ can be achieved.

\begin{figure}[t]
  \centering
  \includegraphics[width=8cm,height=6.5cm]{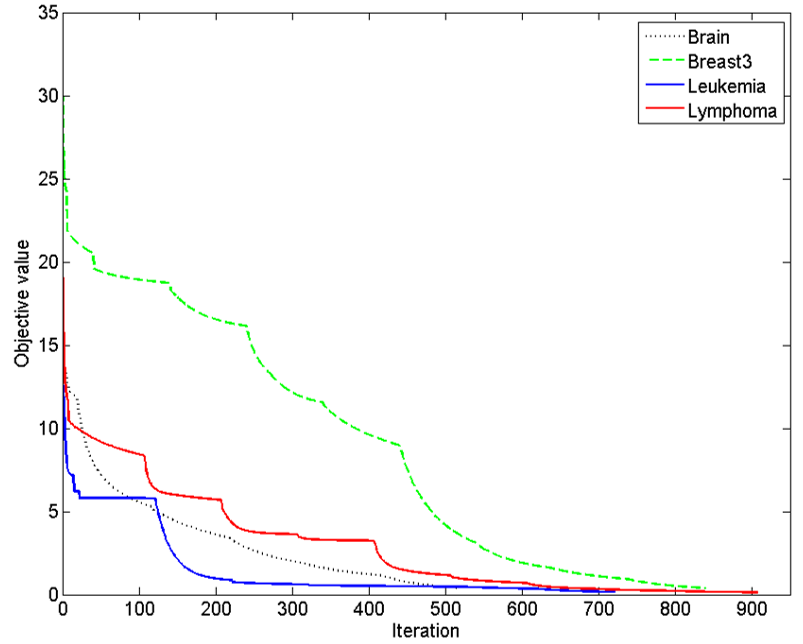}\\
  \caption{An example of objective value at each iteration}\label{objective-value}
\end{figure}

Since the $\lambda$ is monotone decreased, and $\textbf{W}_{\lambda_t}^*$ is set as the initial solution for HIHT in $\lambda_{t+1}$, we obtain that:
\begin{equation}\label{EQ5}
  {\varphi}_{\lambda_t}(\textbf{W}_{\lambda_t}^*) > {\varphi}_{\lambda_{t+1}}(\textbf{W}_{\lambda_t}^*) = {\varphi}_{\lambda_{t+1}}(\textbf{W}_{\lambda_{t+1}}^0) \ge {\varphi}_{\lambda_{t+1}}(\textbf{W}_{\lambda_{t+1}}^*),
\end{equation}
it implies that the objective value is monotone decreasing and a local optimal solution can be achieved by the proposed algorithm. We show an example of the objective value at each iteration in Fig.~\ref{objective-value}, it can be seen that the objective function values monotonically decrease at each iteration until convergence, which verifies the convergence of Alg.~\ref{Alg:HIHT} experimentally.

\subsection{Acceleration of HIHT}
Inspired by \cite{Dong:18}, we derived an acceleration version for HIHT to reduce the computational time. For each fixed value of $\lambda$, Alg.~\ref{Alg:HIHT} iterates to get a solution of problem \eqref{l-202} between steps 6-14. In the following, for acceleration of Alg.~\ref{Alg:HIHT}, we replace steps 6-14 by just calling one outer loop. The outline of the AHIHT method is described as Alg.~\ref{Alg:AHIHT}.
\begin{algorithm}[t]
\caption{Acceleration of HIHT}
\label{Alg:AHIHT}
\begin{algorithmic}
  \STATE \text{\textbf{(Input:)} Training data $\textbf{X} \in R^{d \times N}$, training labels $\textbf{Y} \in R^{C \times N} $;}
  \STATE \text{centering matrix $\textbf{H} \in R^{N \times N}$, parameters $L_0,{\lambda}_0$;}
  \STATE \text{\textbf{(Output:)} \ $\textbf{W}^*$;}
  \STATE \text{1: initialize $k \gets 0, \rho \in (0,1),\gamma >1, \eta >0, \textbf{W}^0=\textbf{0}$;} \\
          \text{2: \textbf{repeat}}\\
          \text{ \ \ \emph{An L-tuning iteration}}\\
          \text{3: \ \ update $\textbf{W}^{k+1}$ by Eq.~\eqref{eq:hiht_solution};}\\
          \text{4: \ \ \textbf{while} \ $ {\varphi}_{\lambda _k}(\textbf{W}^{k})- {\varphi}_{\lambda _k}(\textbf{W}^{k+1})< \frac{\eta}{2} ||\textbf{W}^{k}-\textbf{W}^{k+1}||_F^2$ \ \textbf{do}}\\
          \text{5: \ \ \  $L_{k} \gets \gamma L_{k}$;}\\
          \text{6: \ \ \  update $\textbf{W}^{k+1}$ by Eq.~\eqref{eq:hiht_solution};}\\
          \text{7: \ \ \textbf{end while}}\\
          \text{8: \ \  $L_{k+1} \gets L_{k}$;}\\
          \text{9: \ \ ${\lambda}_{k+1} \gets \rho {\lambda}_k$;}\\
          \text{10: \ \ $k \gets k+1$;}\\
          \text{11: \textbf{until} ${\lambda}_{k+1}$ is small enough}\\
          \text{12: $\textbf{W}^{\star} \gets \textbf{W}^k$.}
\end{algorithmic}
\end{algorithm}
\section{Experiment}\label{Exp}
The proposed $l_{2,0}$-norm regularization multi-class feature selection method is evaluated by several experiments. The experiments are divided into three parts: (1) We evaluate the proposed method in terms of No.fea and ACC using KNN and softmax classifiers, and compare the results with Baseline and other six state-of-the-art feature selection methods. (2) We evaluate the performance sensitivity to the regularization factor $\lambda$. The influence of initialization of AHIHT for feature selection is also evaluated. (3) We compare the convergence speed and computational time of some sparsity-based methods.
\subsection{Data Sets Description}
We use eight biological benchmark datasets to validate the performance of our method in the experiments: Brain \cite{Pomeroy:02}, Breast3 \cite{Veer:02}, Leukemia \cite{Nutt:03}, Lung \cite{Bhattacharjee:01}, Lymphoma \cite{Alizadeh:00}, NCI \cite{Jeffrey:02}, Prostate \cite{Singh:02}, and Srbct \cite{Khan:01}. The $8$ benchmark data sets were downloaded from Feiping Nie's homepage \footnote{http://www.escience.cn/system/file?fileId=82035}, Tab.~\ref{Datasets} gives a detail introduction to these data sets, it can be seen that these data sets are all high-dimensionality with small sample size, thus suitable for evaluating the feature selection task.

\begin{table}[htp]
\caption{Datasets Desciption}\label{Datasets}
\centering
\begin{tabular}{cccc}
\hline\noalign{\smallskip}
{Datasets}&{\#samples}&{\#Features}&{\#Classes}\\
\noalign{\smallskip}\hline\noalign{\smallskip}
{Brain}&{42}&{5597}&{5}\\
{Breast3}&{95}&{4869}&{3}\\
{Leukemia}&{38}&{3051}&{2}\\
{Lung}&{203}&{3312}&{5}\\
{Lymphoma}&{62}&{4026}&{3}\\
{NCI}&{61}&{5244}&{8}\\
{Prostate}&{102}&{6033}&{2}\\
{Srbct}&{63}&{2308}&{4}\\
\noalign{\smallskip}\hline
\end{tabular}
\end{table}
\subsection{Experiment Setup}
In the experiments, the feature selection performance is evaluated on classification accuracy obtained by two popular classifiers, i.e. $K$ nearest neighbor (KNN) and softmax, we set up KNN with $k = 5$. For each data set, $\frac{2}{3}$ of samples per class are randomly selected for training and the rest samples are responsible for testing, ten repeated trials are carried out and average results are recorded for comparison. We compare our feature selection method with Baseline (without feature selection) and six state-of-the-art feature selection algorithms: (1) two basic filter methods: Relief \cite{Kira:92}, and mRMR \cite{Peng:05}; (2) two $l_{2,1}$-norm based methods: RFS \cite{Nie:10}, and RLSR \cite{Chen:17}; (3) two $l_{2,0}$-norm constrained methods: RPMFS \cite{Cai:13}, and EFSF \cite{Pang:19}. The codes of Relief and mRMR are provided by the FEAST package \cite{Brown:12}, and others can be downloaded from the authors' homepages.

For our method, the parameter $\lambda$ is searched in the grid of $\{10^{-5}, 10^{-4}, ..., 10^{0}\}$, and for RFS and RLSR it is searched in $\{10^{-5}, 10^{-3}, ..., 10^{5}\}$. For all methods, the number of selected features is tuned from $\{20, 40, ..., 400\}$ in this paper. All other parameters take the default values as suggested by the authors. For RLSR, all training data are used as labeled data thus it can be seen as a supervised method here. The number of selected features with highest classification accuracy are recorded.
\subsection{Classification Performance}
We evaluate the classification performance of our algorithm in this part. Tab.~\ref{tab:Accuracy} shows the best ACC of each method with corresponding No.fea. From this table, it can be seen that for most datasets, our method can achieve highest ACC with fewest No.fea, or comparable results to the best ones. On NCI dataset, the ACC of our approach obtained by softmax is $78.33\%$, which is $3.33\%$ higher than the second one (expect Baseline). RPMFS, ESFS and our algorithm both solve the original $l_{2,0}$-norm problem, but RPMFS and ESFS try to solve a equality constraint problem so that the number of selected features need to be tuned carefully to obtain a satisfactory classification result. From the numerical comparison, we can see that our method outperforms RPMFS and ESFS most of the time, which means that our method is more efficient than RPMFS and ESFS. The classification results demonstrates that our method can remove more redundant features while maintains the discriminative performance.
\begin{table*}[t]
\footnotesize
  \centering
  \caption{The ACC Value with Corresponding No.fea Using Selected Features (red is the best result and blue is the second one)}
    \begin{tabular}{|p{1.1cm}|p{0.75cm}|p{0.57cm}|p{0.52cm}|p{0.57cm}|p{0.55cm}|p{0.57cm}|p{0.58cm}|p{0.57cm}|p{0.58cm}|p{0.57cm}|p{0.52cm}|p{0.57cm}|p{0.58cm}|p{0.57cm}|p{0.58cm}|}
    \hline
    \multirow{3}[4]{*}{Dataset} & \multicolumn{15}{c|}{KNN} \\
\cline{2-16} & {Baseline} & \multicolumn{2}{c|}{Relief} & \multicolumn{2}{c|}{mRMR} & \multicolumn{2}{c|}{RFS} & \multicolumn{2}{c|}{RLSR} & \multicolumn{2}{c|}{RPMFS} & \multicolumn{2}{c|}{ESFS} & \multicolumn{2}{c|}{$l_{2,0}$-FS} \\
\cline{2-16}  & ACC & No.fea & ACC & No.fea & ACC & No.fea & ACC & No.fea & ACC & No.fea & ACC & No.fea & ACC & No.fea & ACC\\
   \hline
     Brain & 71.67 & 400 & 85.00 & 200 & 81.67 & \textcolor[rgb]{0, 0, 1}{60} & \textcolor[rgb]{ 0, 0, 1}{87.50} & 380 & 80.00 & 340 & 82.50 & 200 & 81.67 & \textcolor[rgb]{1, 0, 0}{100} & \textcolor[rgb]{1, 0, 0}{88.33} \\
    \hline
    Breast3 & 50.65 & 160 & 54.58 & \textcolor[rgb]{0, 0, 1}{380} & \textcolor[rgb]{0, 0,  1}{58.39} & 340 & 55.48 & 340 & 58.06 & 340 & 53.55 & \textcolor[rgb]{0, 0, 1}{380} & \textcolor[rgb]{ 0, 0, 1}{58.39} & \textcolor[rgb]{1, 0, 0}{140} & \textcolor[rgb]{1, 0, 0}{61.29} \\
    \hline
    Leukemia & 96.67 & 400 & 97.50 & 240 & 99.17 & \textcolor[rgb]{0, 0, 1}{260} & \textcolor[rgb]{0, 0, 1}{100.00} & 400 & 97.50 & 400 & 99.17 & 200 & 99.17 & \textcolor[rgb]{1, 0, 0}{80} & \textcolor[rgb]{1, 0, 0}{100.00} \\
    \hline
    Lung & 94.39 & \textcolor[rgb]{0.00,0.00,1.00}{380} & \textcolor[rgb]{0.00,0.00,1.00}{94.39} & 200 & 94.09 & \textcolor[rgb]{1, 0, 0}{80} & \textcolor[rgb]{1, 0, 0}{95.67} & 400 & 91.67 & 300 & 92.88 & 400 & 91.82 & {200} & {93.33} \\
    \hline
    Lymphoma & 97.50 & 340 & 99.50 & \textcolor[rgb]{1, 0, 0}{60} & \textcolor[rgb]{1, 0, 0}{100.00} & 140 & 100.00 & 260 & 99.00 & 280 & 99.00 & 20 & 99.00 & \textcolor[rgb]{0, 0, 1}{40} & \textcolor[rgb]{0, 0, 1}{100.00} \\
    \hline
    NCI & 73.89 & 340 & 73.33 & \textcolor[rgb]{0, 0, 1}{240} & \textcolor[rgb]{0, 0, 1}{73.89} & \textcolor[rgb]{0, 0, 1}{240} & \textcolor[rgb]{0, 0, 1}{73.89} & 220 & 72.78 & 380 & 67.78 & 320 & 73.33 & \textcolor[rgb]{1, 0, 0}{60} & \textcolor[rgb]{ 1, 0, 0}{74.44} \\
    \hline
    Prostate & 81.82 & \textcolor[rgb]{0, 0, 1}{40} & \textcolor[rgb]{0, 0, 1}{91.82} & 60 & 90.30  & \textcolor[rgb]{1, 0, 0}{20} & \textcolor[rgb]{1, 0, 0}{93.03} & 20 & 89.09 & 20 & 86.06 & 20 & 87.88 & \textcolor[rgb]{1, 0, 0}{20} & \textcolor[rgb]{1, 0, 0}{93.94} \\
    \hline
    Srbct & 93.16 & 240 & 98.95 & 260 & 98.95 & \textcolor[rgb]{1, 0, 0}{20} & \textcolor[rgb]{1,  0, 0}{100.00} & 60 & 96.84 & 380 & 93.68 & 400 & 100.00 & \textcolor[rgb]{0, 0, 1}{40} & \textcolor[rgb]{0, 0, 1}{100.00} \\
    \hline
    \multirow{3}[4]{*}{Dataset} & \multicolumn{15}{c|}{Softmax} \\
\cline{2-16} & {Baseline} & \multicolumn{2}{c|}{Relief} & \multicolumn{2}{c|}{mRMR} & \multicolumn{2}{c|}{RFS} & \multicolumn{2}{c|}{RLSR} & \multicolumn{2}{c|}{RPMFS} & \multicolumn{2}{c|}{ESFS} & \multicolumn{2}{c|}{$l_{2,0}$-FS} \\
\cline{2-16}  & ACC & No.fea & ACC& No.fea & ACC & No.fea & ACC & No.fea & ACC & No.fea & ACC & No.fea & ACC & No.fea & ACC\\
   \hline
  Brain & 82.50 & 400 & 88.33 & 400 & 88.33 & \textcolor[rgb]{0, 0, 1}{160} & \textcolor[rgb]{0,  0, 1}{92.50} & 220 & 85.83 & 280 & 88.33 & 400 & 85.42 & \textcolor[rgb]{1, 0, 0}{180} & \textcolor[rgb]{1, 0, 0}{93.33}\\
    \hline
    Breast3 & 58.71 & 280 & 59.35 & 240 & 58.06 & 340 & 61.94 & \textcolor[rgb]{0, 0, 1}{300} & \textcolor[rgb]{ 0, 0, 1}{61.94} & 340 & 60.97 & 400 & 58.39 & \textcolor[rgb]{1, 0, 0}{280} & \textcolor[rgb]{1, 0, 0}{62.90} \\
    \hline
    Leukemia & 99.17 & 320  & 98.33 & 60 & 99.17 & 120 & 99.17 & \textcolor[rgb]{0, 0, 1}{220} & \textcolor[rgb]{0, 0, 1}{100.00} & 240 & 100.00 & 400 & 99.17 & \textcolor[rgb]{1, 0, 0}{120} & \textcolor[rgb]{1, 0, 0}{100.00} \\
    \hline
    Lung & \textcolor[rgb]{0.00,0.00,1.00}{95.76} & 280 & 94.70 & 400 & 95.30 & \textcolor[rgb]{1, 0, 0}{180} & \textcolor[rgb]{1,  0, 0}{96.36} & 320 & 93.94 & 340 & 94.55 & 400 & 94.39 &{300} & {95.61} \\
    \hline
   Lymphoma & 94.00 & 220 & 95.00 & 20 & 95.50 & \textcolor[rgb]{0, 0, 1}{60} & \textcolor[rgb]{0, 0,  1}{98.50} & 60  & 96.00 & 320 & 97.50 & \textcolor[rgb]{0, 0, 1}{60} & \textcolor[rgb]{0, 0,  1}{98.50} & \textcolor[rgb]{1, 0, 0}{280} & \textcolor[rgb]{1, 0, 0}{99.50} \\
    \hline
    NCI & \textcolor[rgb]{0, 0, 1}{76.11} & 400 & 75.00 & 140 & 72.22 & 140 & 73.33 & 360 & 71.67 & 400 & 72.22 & 260 & 73.33 & \textcolor[rgb]{1, 0, 0}{220} & \textcolor[rgb]{1, 0, 0}{78.33} \\
    \hline
   Prostate & 90.61 & 120 & 92.73 & 260 & 92.12 & \textcolor[rgb]{0, 0, 1}{140} & \textcolor[rgb]{ 0, 0, 1}{95.45} & 400 & 93.94 & 280 & 91.52 & 260 & 93.33 & \textcolor[rgb]{1, 0, 0}{100} & \textcolor[rgb]{1, 0, 0}{95.76} \\
    \hline
    Srbct & 99.47 & 240 & 98.42 & 20  & 97.89 & \textcolor[rgb]{1, 0, 0}{20} & \textcolor[rgb]{1,  0, 0}{100.00} & 160 & 97.37 & 180 & 96.84 & 280 & 98.95 & \textcolor[rgb]{0, 0, 1}{80} & \textcolor[rgb]{0, 0, 1}{100.00} \\
    \hline
    \end{tabular}%
  \label{tab:Accuracy}%
\end{table*}%

\begin{figure*}[htp]
  \begin{center}
    \subfigure[]{\label{Fig:Breast3}
        \includegraphics[width=6.5cm,height=5cm]{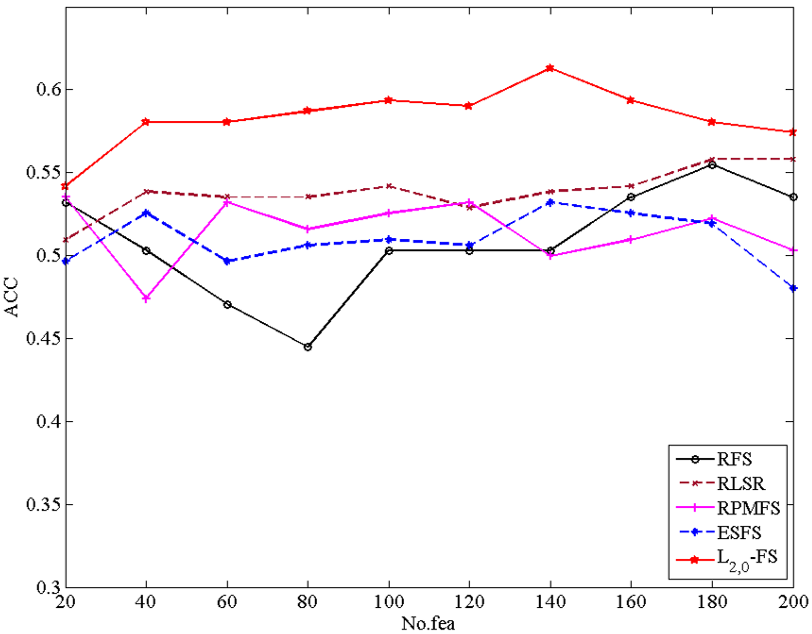}
    }
    \subfigure[]{\label{Fig:Lung}
        \includegraphics[width=6.5cm,height=5cm]{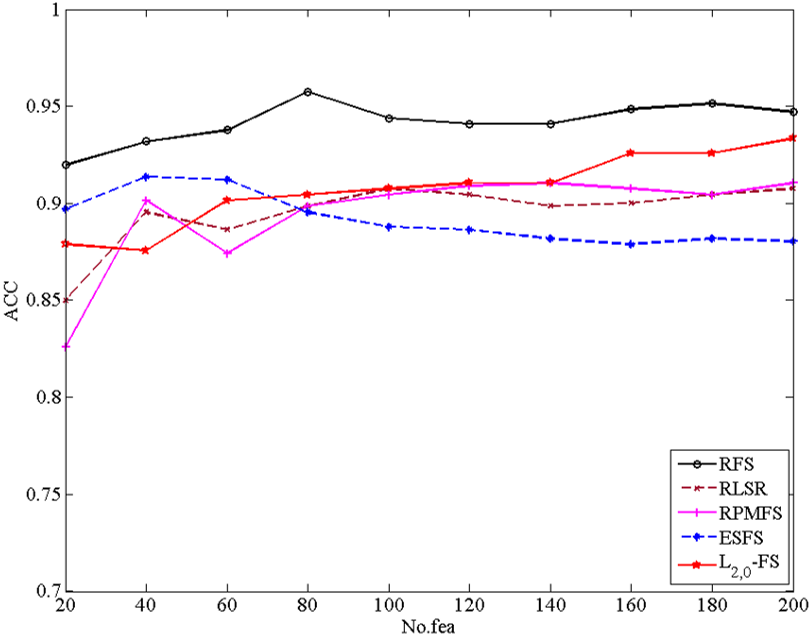}
    }
    \subfigure[]{\label{Fig:NCI}
        \includegraphics[width=6.5cm,height=5cm]{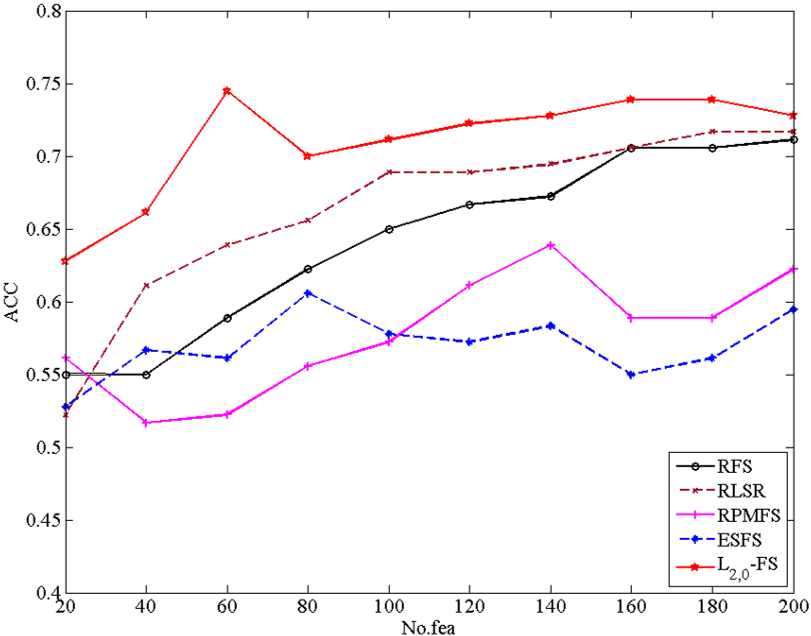}
    }
    \subfigure[]{\label{Fig:Srbct}
        \includegraphics[width=6.5cm,height=5cm]{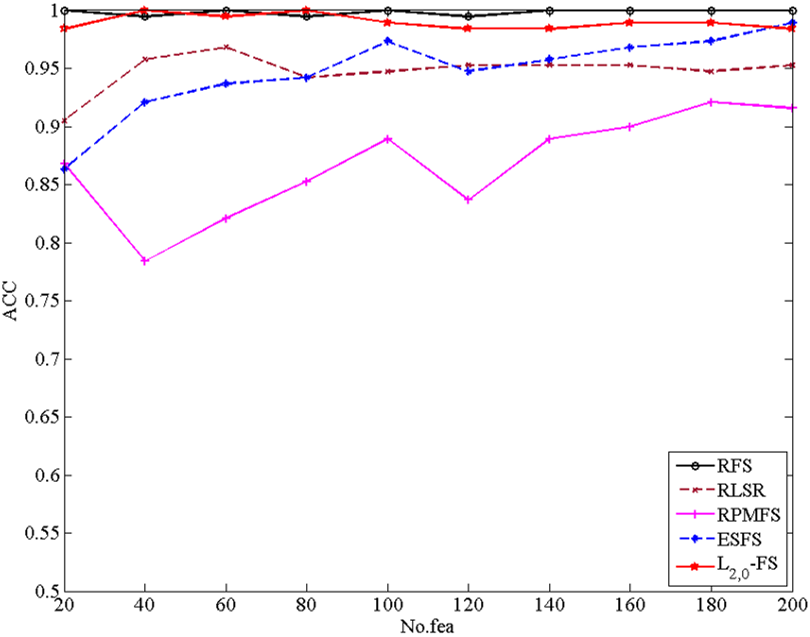}
    }
    \caption{\label{fig:accvsnum1}ACC VS. No.fea using selected features by KNN.
      \subref{Fig:Breast3} Breast3.
      \subref{Fig:Lung} Lung.
      \subref{Fig:NCI} NCI.
      \subref{Fig:Srbct} Srbct.
      }
  \end{center}
\end{figure*}

\begin{figure*}[htp]
  \begin{center}
    \subfigure[]{\label{Fig:Brain}
        \includegraphics[width=6.5cm,height=5cm]{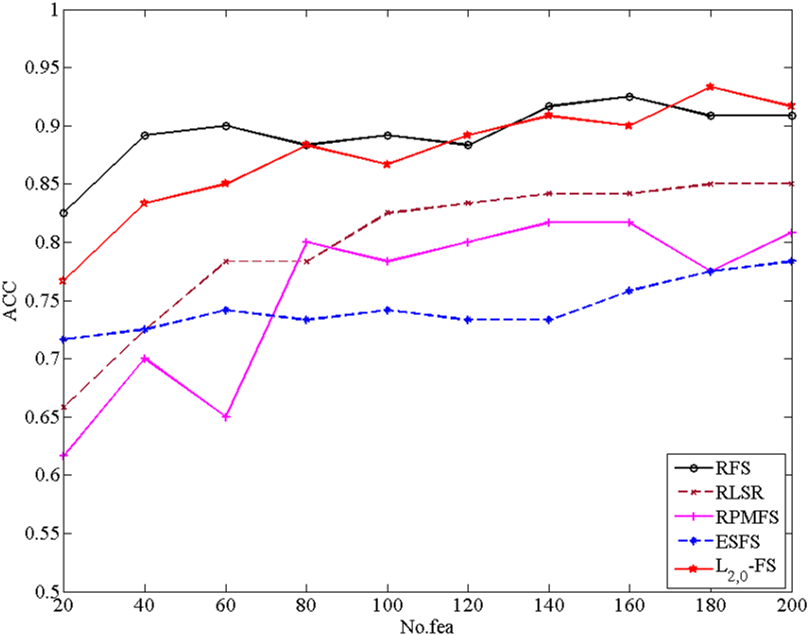}
    }
    \subfigure[]{\label{Fig:Leukemia}
        \includegraphics[width=6.5cm,height=5cm]{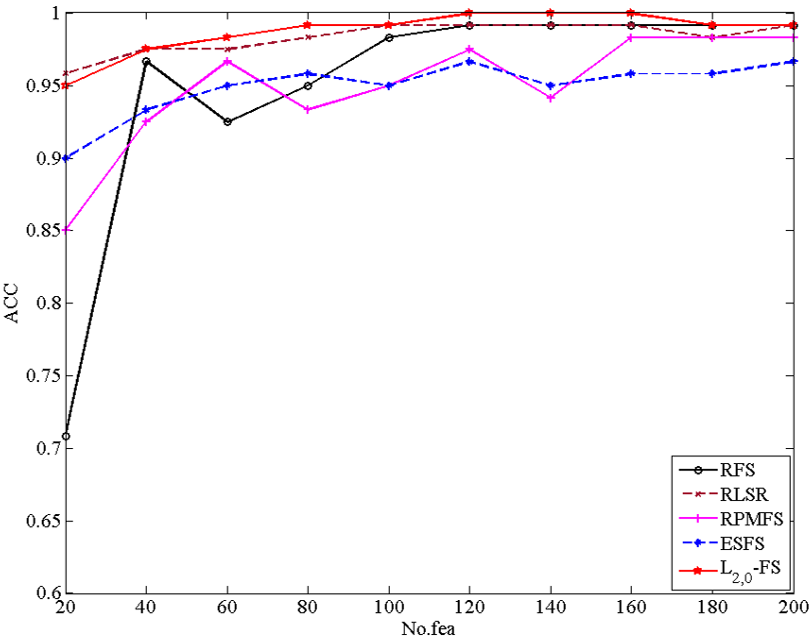}
    }
    \subfigure[]{\label{Fig:Lymphoma}
        \includegraphics[width=6.5cm,height=5cm]{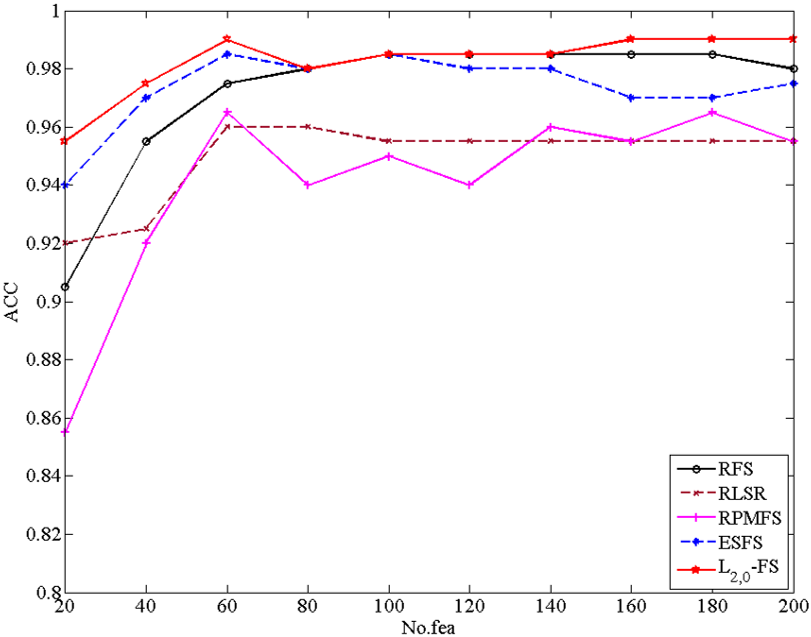}
    }
    \subfigure[]{\label{Fig:Prostate}
        \includegraphics[width=6.5cm,height=5cm]{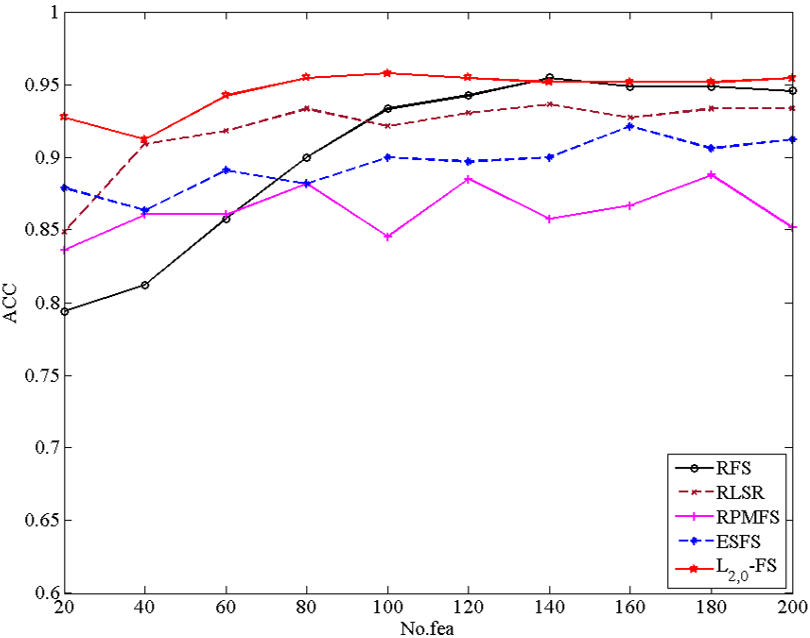}
    }
    \caption{\label{fig:accvsnum2}ACC VS. No.fea using selected features by softmax.
      \subref{Fig:Brain} Brain.
      \subref{Fig:Leukemia} Leukemia.
      \subref{Fig:Lymphoma} Lymphoma.
      \subref{Fig:Prostate} Prostate.
      }
  \end{center}
\end{figure*}

Fig.~\ref{fig:accvsnum1} and Fig.~\ref{fig:accvsnum2} show ACC V.S. No.fea obtained by KNN and softmax, respectively. It is obvious that the proposed method distinctly outperforms other
approaches in the most experimental data sets. What's more, compared with other lines, ours are not fluctuated too much as the No.fea changed, which indicates the performance of the proposed approach is much robust to the number of selected features than other methods.

\subsection{Parameter Sensitivity}
\begin{figure*}[htp]
  \begin{center}
    \subfigure[]{\label{Fig:Brain1}
        \includegraphics[width=7.5cm,height=5.5cm]{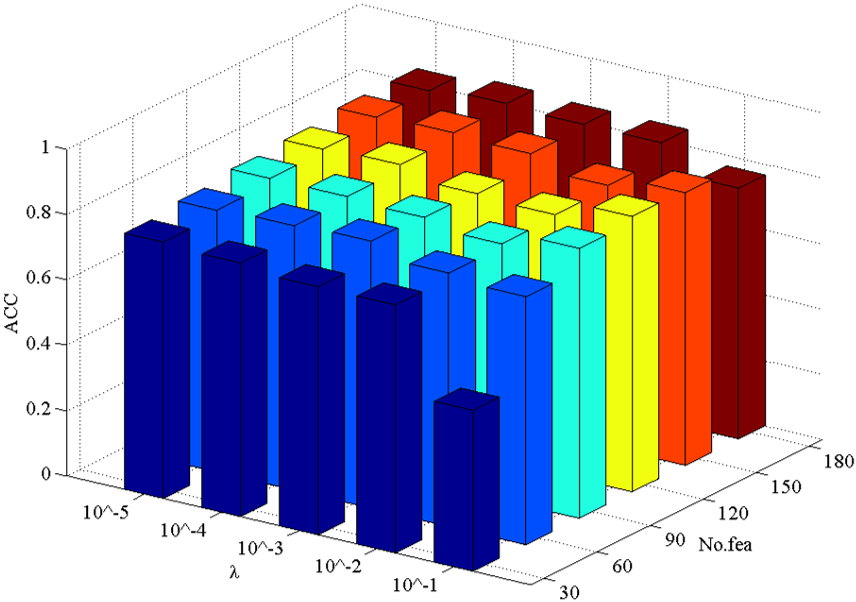}
    }
    \subfigure[]{\label{Fig:Leukemia1}
        \includegraphics[width=7.5cm,height=5.5cm]{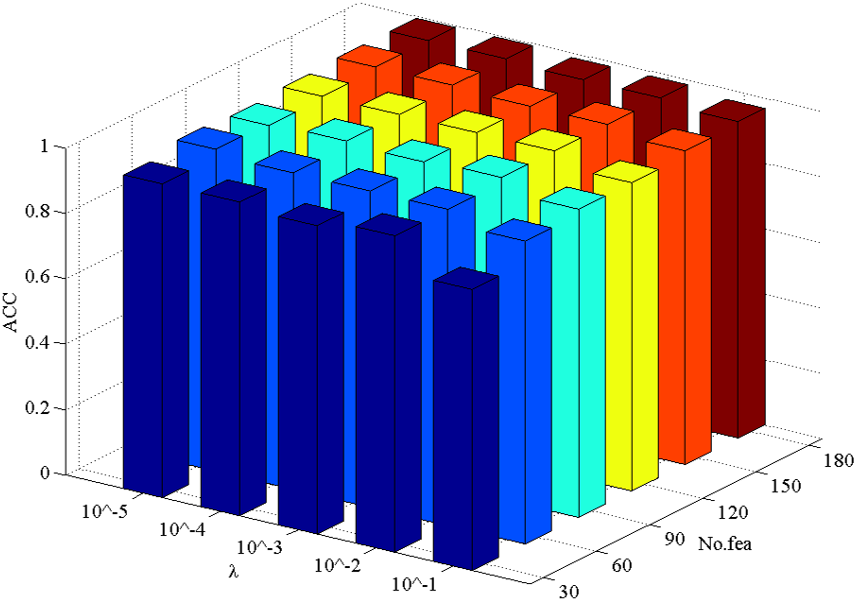}
    }
    \subfigure[]{\label{Fig:Lung1}
        \includegraphics[width=7.5cm,height=5.5cm]{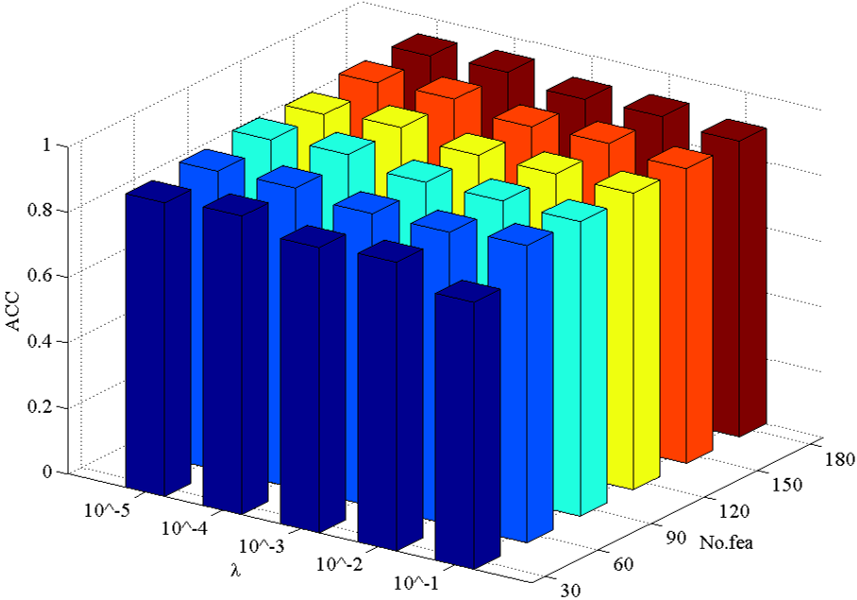}
    }
    \subfigure[]{\label{Fig:Prostate1}
        \includegraphics[width=7.5cm,height=5.5cm]{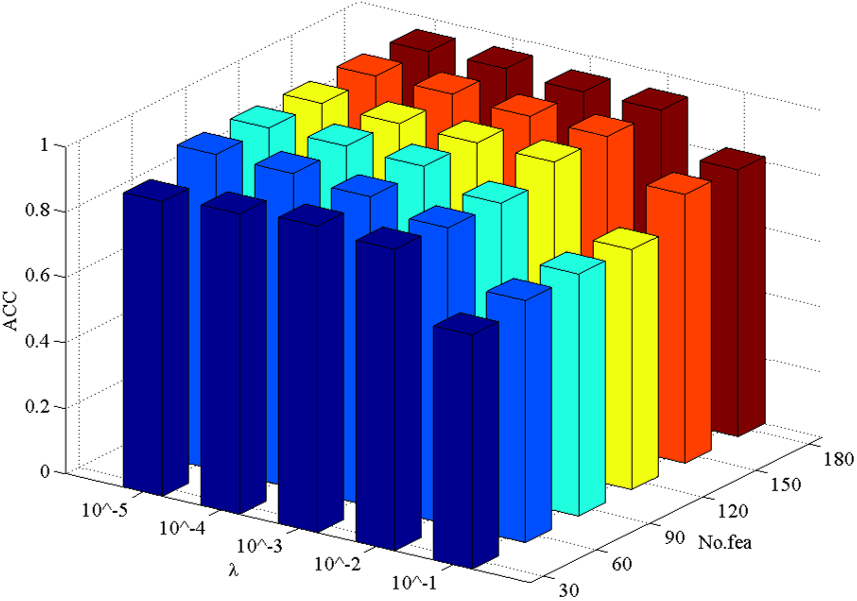}
    }
    \caption{\label{fig:lambda}Parameter Sensitivity demonstration on different data sets.
      \subref{Fig:Brain} Brain.
      \subref{Fig:Leukemia1} Leukemia.
      \subref{Fig:Lung1} Lung.
      \subref{Fig:Prostate1} Prostate.
      }
  \end{center}
\end{figure*}

In this part, we evaluate the sensitivity of $\lambda$, the ACC is employed to evaluate the performance of classification with $\lambda$ searched in the grid of $\{10^{-5}, 10^{-4}, 10^{-3}, 10^{-2}, 10^{-1}\}$. The number of selected features varies in $\{30, 60, ...,
150, 180\}$. The datasets of Brain, Leukemia, Lung and Prostate are used for testing, and the experimental results are shown in Fig.~\ref{fig:lambda}. It can be seen from this figure that, the performance of $l_{2,0}$-FS is not very sensitive to $\lambda$. When $\lambda$ is in the range of $\{10^{-5},...,10^{-2}\}$, there is almost no fluctuation on ACC. Thus the proposed approach is robust to $\lambda$ and it is no need to speed much time to tune the value of $\lambda$. It also can be seen that the performance get worst when $\lambda =10^{-1}$, the reason is that in this case the proposed method will obtain a very sparse solution and the number of selected features tend to zero.

\begin{figure*}[htp]
  \begin{center}
    \subfigure[]{\label{Fig:Breast32}
        \includegraphics[width=5cm,height=4cm]{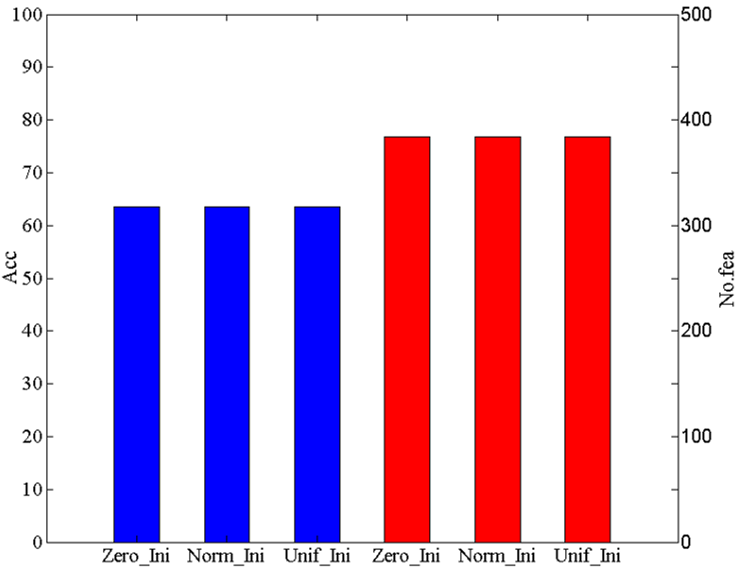}
    }
    \subfigure[]{\label{Fig:Lung2}
        \includegraphics[width=5cm,height=4cm]{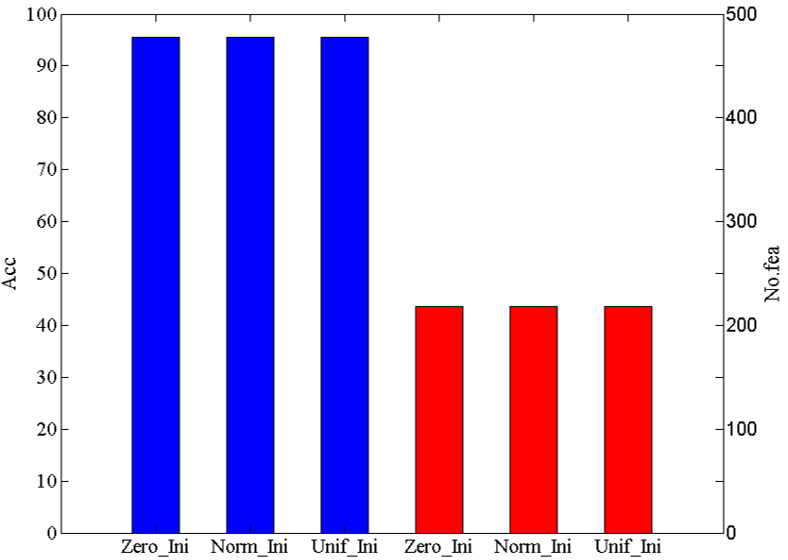}
    }
    \subfigure[]{\label{Fig:Srbct2}
        \includegraphics[width=5cm,height=4cm]{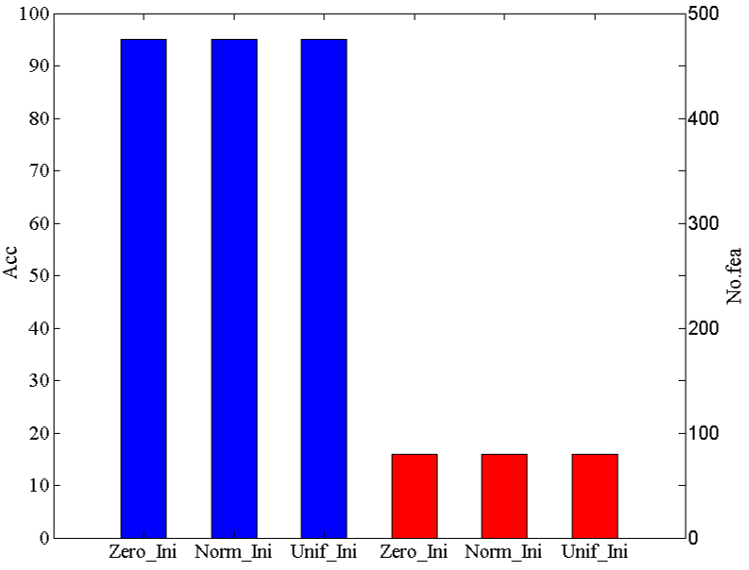}
    }
    \caption{\label{fig:differentinit}Results of classification accuracy and the number of selected features with respect to different initialization.
      \subref{Fig:Breast32}: Breast3.
      \subref{Fig:Lung2}: Lung.
      \subref{Fig:Srbct2}: Srbct.
      }
  \end{center}
\end{figure*}

$L_{2,0}$-norm is a non-convex problem, which may make the solution sensitive to initialization. Three kinds of initialization are used to explore the effect: zero initialization, random Gaussian distribution initialization, and random uniform distribution initialization. For Gaussian and uniform distribution, ten repeated trials are carried out and average results are recorded. Fig.~\ref{fig:differentinit} shows the results (in this figure, the No.fea is equal to the number of non-zero rows of $\textbf{W}$), in which datasets Breast3, Lung, and Srbct are used. It can be seen that these three different initialization methods can get the same results, indicating that AHIHT is not sensitive to initialization when applied to feature selection.
\subsection{Comparison of Convergence Speed and Time Consumption}
\begin{figure}[htp]
\begin{center}
    \subfigure[]{\label{Fig:Leukemia3}
        \includegraphics[width=4cm,height=3.5cm]{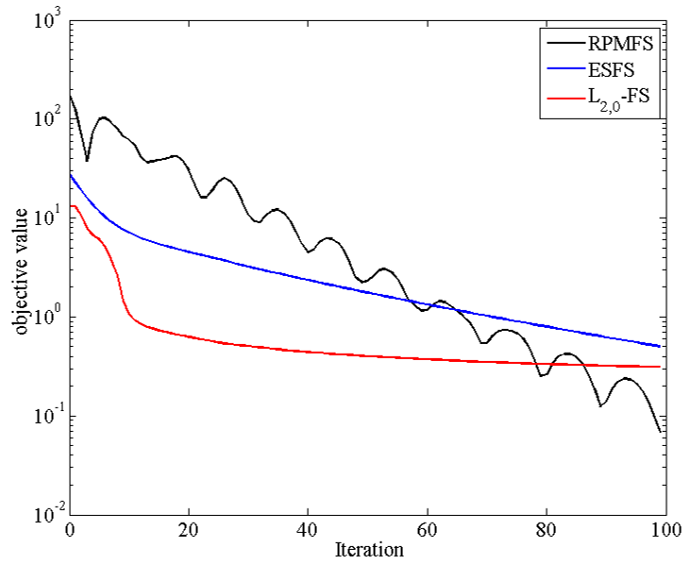}
    }
    \subfigure[]{\label{Fig:Prostate3}
        \includegraphics[width=4cm,height=3.5cm]{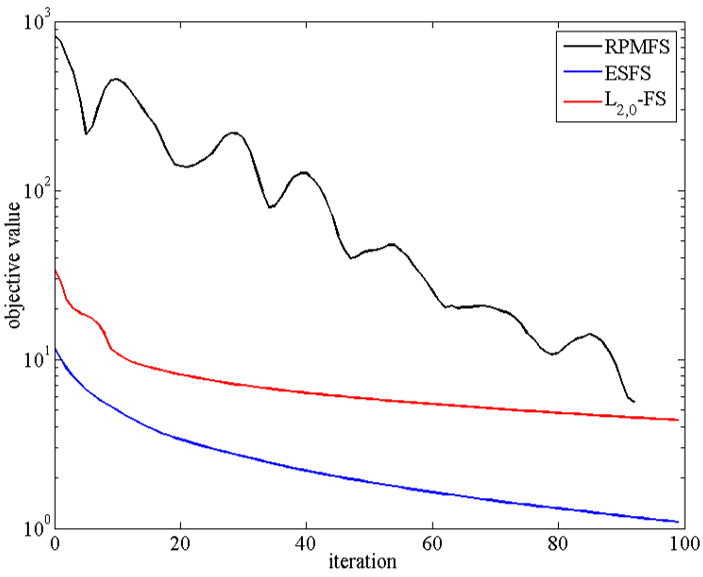}
    }
    \caption{\label{fig:convergence}The objective function value v.s. iteration number.
      \subref{Fig:Leukemia3}: Leukemia.
      \subref{Fig:Prostate3}: Prostate.
      }
  \end{center}
\end{figure}

\begin{figure}[t]
  \centering
  \includegraphics[width=7cm,height=5.5cm]{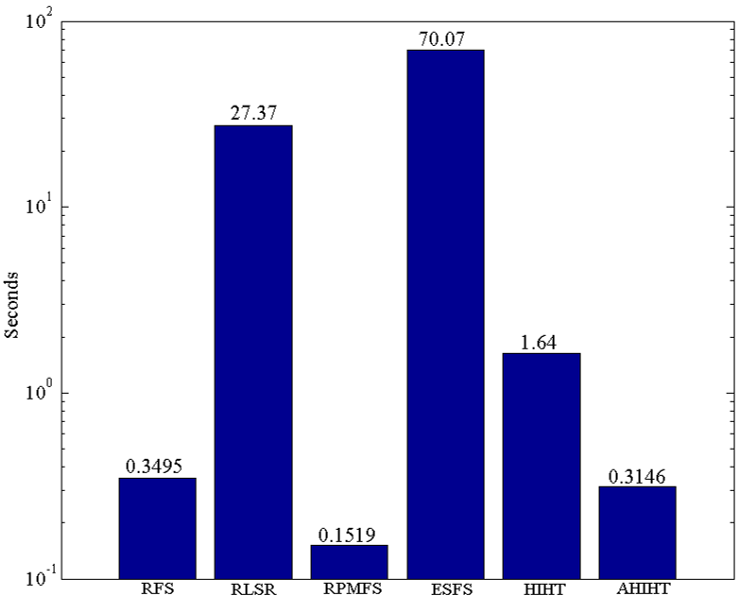}\\
  \caption{The average computational time of each sparsity regularization method}\label{time}
\end{figure}

Fig.~\ref{fig:convergence} plots the objective function value for each iteration of the three $l_{2,0}$-norm based methods. As can be observed, our method can decrease the objective value quickly in early iterations, which indicates that our method converges fast than RPMFS and ESFS. Fig.~\ref{time} shows the average computational time of each sparsity regularization method on the eight datasets, which also compares AHIHT with HIHT. It can be seen that the computational time of RFS, RPMFS and AHIHT are much less than the other three methods, the reason is that the computational complexity of RFS, RPMFS and AHIHT is in propotion to $d$ while for other methods it is in propotion to $d^2$ or $d^3$. Comparing AHIHT with HIHT, it can be found that AHIHT can reduce the computational time of HIHT effectively, thus is more suitable for practical applications.

\subsection{Comparison of AHIHT and HIHT}
\begin{figure*}[htp]
  \begin{center}
    \subfigure[]{\label{Fig:Breast34}
        \includegraphics[width=5cm,height=4cm]{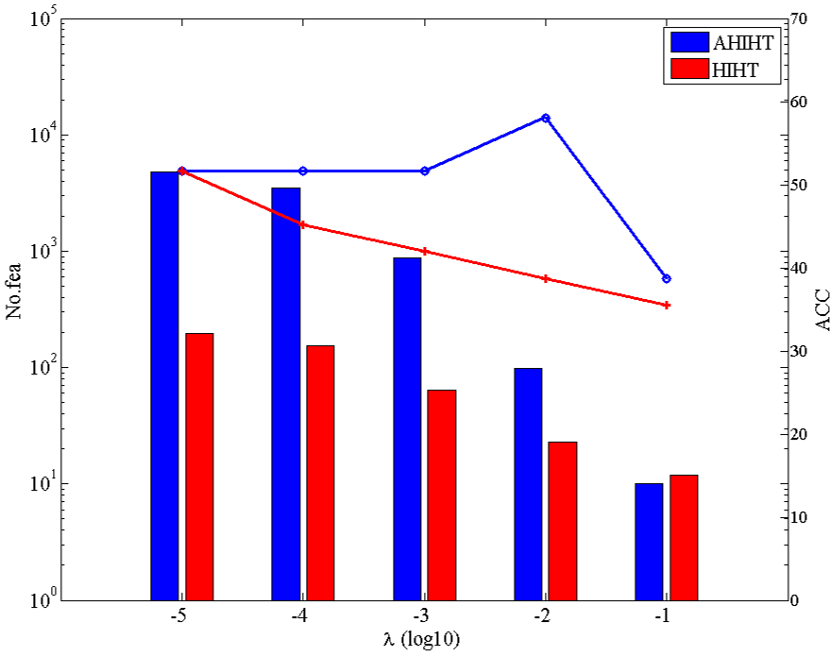}
    }
    \subfigure[]{\label{Fig:Lung4}
        \includegraphics[width=5cm,height=4cm]{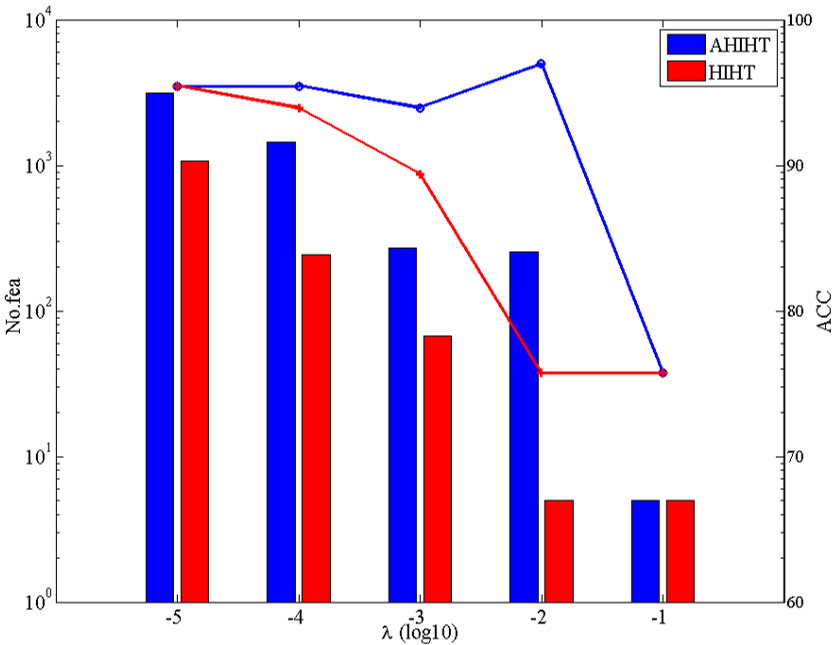}
    }
    \subfigure[]{\label{Fig:NCI4}
        \includegraphics[width=5cm,height=4cm]{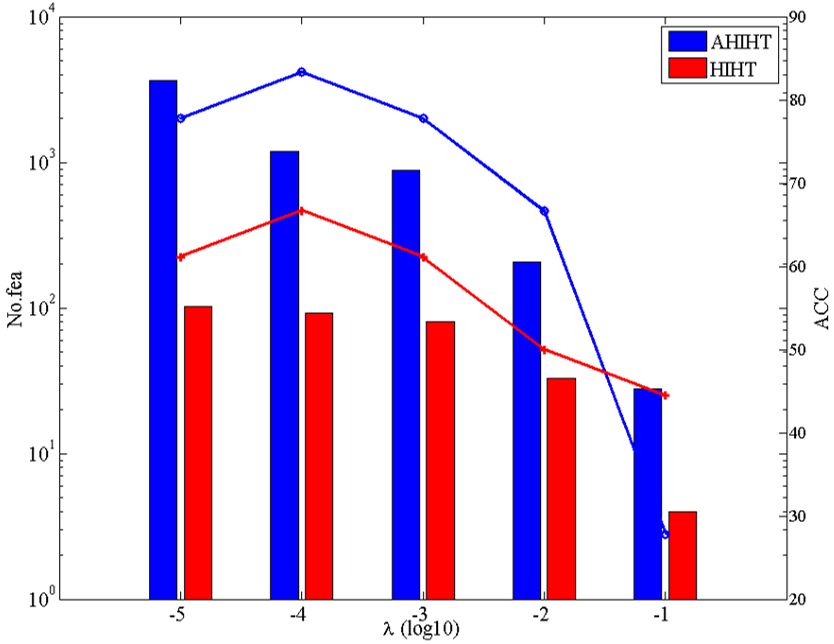}
    }
    \caption{\label{fig:comparisonofHIHTandAHIHT}Classification accuracy and the number of selected features with respect to $\lambda$ of AHIHT and HIHT.
      \subref{Fig:Breast34}: Breast3.
      \subref{Fig:Lung4}: Lung.
      \subref{Fig:NCI4}: NCI.
      }
  \end{center}
\end{figure*}

In this part, we compare the performance of HIHT and its accelerated version in terms of accuracy and the row-sparsity of produced solutions with different values of $\lambda$. Fig.~\ref{fig:comparisonofHIHTandAHIHT} show the results, in which datasets Breast3, Lung, and NCI are used. From this figure it can be seen that, when $\lambda$ is less than $0.1$, HIHT can get a more sparse solution than AHIHT, while AHIHT can get higher classification accuracy than HIHT, it means that the features selected by AHIHT is more useful for classification than those of HIHT. What's more, the results obtained by AHIHT demonstrate AHIHT is more robust to $\lambda$ than HIHT. For HIHT, it should tune a small value of $\lambda$ to get satisfactory result, while a small value of $\lambda$ will spend more computational time, which has been show in Fig.~\ref{time}. In conclusion, AHIHT is more practical than HIHT for feature selection.
\section{Conclusions}\label{Con}
In this paper, we proposed a novel method to solve the original $l_{2,0}$-norm regularization least square problem for multi-class feature selection, instead of solving its relax problem like most of other existing methods. A homotopy iterative hard threshold is proposed to optimize the proposed model which can obtain exact row-sparsity solution. Besides, in order to reduce the computational time of HIHT for feature selection task, an acceleration version of HIHT (AHIHT) is derived. Experiments on eight biological datasets show that we can achieve comparable or better classification performance comparing with other six state-of-the-art feature selection algorithms. In the future, we are interested in combining the $l_{2,1}$-norm loss function with the $l_{2,0}$-norm regularization.

\bibliographystyle{IEEEtran}
\bibliography{references}
\end{document}